\documentclass{article}

\usepackage[preprint,nonatbib]{neurips_data_2022}

\usepackage[utf8]{inputenc} 
\usepackage[T1]{fontenc}    
\usepackage{hyperref}       
\usepackage{url}            
\usepackage{booktabs}       
\usepackage{amsfonts}       
\usepackage{nicefrac}       
\usepackage{microtype}      
\usepackage{xcolor}         
\usepackage{amsmath}
\usepackage{graphics}
\usepackage[dvipdfmx]{graphicx}
\usepackage {arydshln}
\usepackage[subrefformat=parens]{subcaption}
\usepackage{fancybox}
\usepackage{ascmac}

\newcommand{\Hline}{\noalign{\hrule height 1.0pt}} 
\usepackage{pifont}
\newcommand{\cmark}{\ding{51}}%
\newcommand{\xmark}{\ding{55}}%

\title{An Empirical Study of Personalized Federated Learning}

\author{%
  Koji Matsuda, Yuya Sasaki, Chuan Xiao, Makoto Onizuka \\
  Graduate School of Information Science and Technology\\
  Osaka University\\
  \texttt{\{matsuda.koji, sasaki, chuanx, onizuka\} @ist.osaka-u.ac.jp} \\
}

\begin{document}

\maketitle

\begin{abstract}

Federated learning is a distributed machine learning approach in which a 
single server and multiple clients collaboratively build machine learning 
models without sharing datasets on clients. A challenging issue of 
federated learning is data heterogeneity (i.e., data distributions may 
differ across clients). To cope with this issue, numerous federated learning 
methods aim at personalized federated learning and build optimized models 
for clients. Whereas existing studies empirically evaluated their own methods, 
the experimental settings (e.g., comparison methods, datasets, and client 
setting) in these studies differ from each other, and it is unclear which 
personalized federate learning method achieves the best performance and how 
much progress can be made by using these methods instead of standard (i.e., 
non-personalized) federated learning.
In this paper, we benchmark the performance of existing personalized federated 
learning through comprehensive experiments to evaluate the characteristics of 
each method. Our experimental study shows that (1) there are no champion 
methods, (2) large data heterogeneity often leads to high accurate predictions, 
and (3) standard federated learning methods (e.g. FedAvg) with fine-tuning 
often outperform personalized federated learning methods. We open our benchmark 
tool {\sf FedBench} for researchers to conduct experimental studies with various 
experimental settings.

\end{abstract}

\section{Introduction}
\label{sec:1_introduction}

Federated learning has emerged as a distributed machine learning approach 
in which a single server and multiple clients collaboratively build machine 
learning models without sharing datasets on clients in order to reduce  
privacy risks and communication traffic~\cite{mcmahan2017communication}.
Its general procedure consists of two steps: ($1$) client training, in which 
clients train models on their local data and send their trained models to 
the server, and ($2$) model aggregation, in which the server aggregates those 
models to build a global model and distributes the global model to the clients. 
Due to its effectiveness in a distributed scenario, federated learning has 
received considerable attention from research communities and numerous methods 
have been proposed~\cite{chen2021fedbe,dai2020federated,li2020practical,he2020group,karimireddy2020scaffold,MLSYS2020_38af8613,Wang2020Federated,fednova,yurochkin2019bayesian}.



One of the main challenges in federated learning lies on data heterogeneity: 
clients have local data that differ in distributions', i.e., they do not conform 
to the property of independent and identically distributed (IID) random 
variables. This causes difficulty in learning a single global model that is 
optimal for all clients. It was reported that, in typical federated learning 
methods, model parameters of a global model are divergent when each client has 
non-IID local data~\cite{MLSYS2020_38af8613,li2019convergence}. To deal with 
this issue, a recent trend resorts to {\it personalized federated learning}, 
which aims to build {\it personalized models} that are optimized for 
clients~\cite{FedPer,FedRep,Ditto,lgfedavg,mansour2020three,FedMe,shen2020federated,NEURIPS2020_f4f1f13c,fedfomo}.

\noindent
{\bf Motivation.}
While empirical evaluations are available in existing studies, the experimental 
settings (e.g., comparison methods, datasets, and client settings) in these 
studies differ from each other. Despite a comprehensive comparison and analysis 
of non-personalized federated learning methods on non-IID data~\cite{li2022federated}, 
to the best of our knowledge, a comprehensive comparison and analysis of 
personalized federated learning methods have not been conducted yet. 
Hence the following three questions remain unanswered: 
\begin{itemize}
    \item Which personalized federated learning method performs the best?
    \item Do personalized federated learning methods perform better than 
    non-personalized federated learning methods?    
    \item How does the experimental setting affect the model performance?
\end{itemize}

\noindent
{\bf Contributions.}
In this paper, we benchmark the performance of various personalized federated 
learning methods in comprehensive experimental studies. 
For future method development, we summarize several key findings of our study:
\begin{itemize}
    \item {\bf There are no champion methods}: none of the existing 
    state-of-the-art personalized federated learning methods outperform the 
    others in all the cases. 
    \item {\bf Fine-tuning works well for data heterogeneity}: standard 
    federated learning methods with fine-tuning are able to build highly 
    accurate personalized models, which are not evaluated fairly in existing 
    studies.    
    \item {\bf Large data heterogeneity often leads to high accuracy}: the 
    larger the degree of data heterogeneity, the more accurate the personalized 
    federated learning methods are.
\end{itemize}

To foster future work, we develop {\sf FedBench}, a Jupyter notebook-based tool, 
which supports performing easily experimental studies with various methods, 
experimental settings, and datasets. {\sf FedBench} is publicly available at 
\url{https://github.com/OnizukaLab/FedBench} under the MIT licence.


\section{Preliminaries on Personalized Federated Learning}
\label{sec:2_relatedwork}

\subsection{Problem Formulation}
We describe the problem formulation of personalized federated learning. 
Consider a server and a set of clients which collaboratively build 
personalized models of clients. Let $S$ denote the set of clients. 
$|S|$ is the number of clients. We use a subscript $i$ for the index of 
the $i$-th client. $D_i$ denotes the local data of client $i$. $n_i$ 
denotes the number of data samples (e.g., records, images, and 
texts). $N$ denotes the sum of $n_i$ across all the clients. $x_i$ and 
$y_i$ are the features and the labels of samples contained in the local 
data of client $i$, respectively. $T$ and $E$ are the total numbers of 
global communication rounds and local training rounds, respectively, 
where global communication refers to the communication between the 
server and the clients during training and local training refers to the 
training of each client's model using its local data. 

In standard federated learning, a server and clients aim to create a 
single global model $w_g$. We define standard federated learning as 
the following optimization problem:
\begin{gather}
\underset{w_g\in \mathbb{R}^d}{\min}\sum\limits^{|S|}_{i=1}\mathcal{T}_i(w_g), 
\end{gather}
where $\mathcal{T}_i$ is the objective for client $i$ and is defined as follows:
\begin{gather}
    \mathcal{T}_i (w)=\frac{1}{n_i}\sum\limits_{(x_i,y_i)\in D_i}f_i(x_i,y_i,w), 
\end{gather}
where $f_i$ is a loss function. 

In personalized federated learning, a server and clients aim to create 
a personalized model $w_p$ for each client. We define personalized 
federated learning as the following optimization problem: 
\begin{gather}
\underset{\{w_{p_1},\ldots,w_{p_{|S|}}\}\in \mathbb{R}^d}{\min}\sum\limits^{|S|}_{i=1}\mathcal{T}_i(w_{p_i}), 
\end{gather}
where $w_{{p}_i}$ is the personalized model of client $i$.

\subsection{Representative Federated Learning Methods}
We introduce a set of representative federated learning methods which are 
evaluated in this paper. The characteristics are summarized in Table~\ref{tab:method}.

\begin{table}[!t]
\centering
\caption{Summary of federated learning methods. Model splitting indicates 
that the server and each client train a part of the model. Model update 
regularization is a technique by which the personalized model trained by 
each client will not stray too far from the global model.}
\label{tab:method}
\scalebox{0.69}{
\begin{tabular}{lcccccccccc}\Hline
Method & FedAvg & FedProx & HypCluster & FML & FedMe & LG-FedAvg & FedPer & FedRep & Ditto & pFedMe \\\hline
Personalization & \xmark & \xmark & \cmark & \cmark & \cmark & \cmark & \cmark & \cmark & \cmark & \cmark \\
Clustering & \xmark & \xmark & \cmark & \xmark & \cmark & \xmark & \xmark & \xmark & \xmark & \xmark \\
Deep mutual learning & \xmark & \xmark & \xmark & \cmark & \cmark & \xmark & \xmark & \xmark & \xmark & \xmark \\
Model splitting & \xmark & \xmark & \xmark & \xmark & \xmark & \cmark & \cmark & \cmark & \xmark & \xmark \\
Model update regularization& \xmark & \cmark & \xmark & \xmark & \xmark & \xmark & \xmark & \xmark & \cmark & \cmark \\
\Hline
\end{tabular}
}
\end{table}

\textbf{Standard federated learning.}
The basic method on federated learning is FedAvg~\cite{mcmahan2017communication}, 
which aggregates all the trained models of the clients by averaging their 
model parameters to build a single global model. 
FedProx~\cite{MLSYS2020_38af8613} utilizes a regularization term in its loss 
function so that the clients' trained models will not significantly differ from 
the global model.


\textbf{Personalized federated learning.}
Hypcluster~\cite{mansour2020three} is a method that divides the set of clients 
into groups and creates a model for each group. Federated mutual learning 
(FML)~\cite{shen2020federated} and FedMe~\cite{FedMe} use deep mutual 
learning~\cite{8578552}, which is a machine learning method by which multiple 
models are trained to imitate each other's output. In FML, each client trains 
its own personalized model independently and a generalized model is trained 
collaboratively. In FedMe, each client trains its own personalized model and 
other clients' personalized models, depending on a clustering. In 
LG-FedAvg~\cite{lgfedavg}, FedPer~\cite{FedPer}, and FedRep~\cite{FedRep}, the 
server and each client train a part of the model. These methods combine the 
server's and the clients' sub-models for training and inference. In LG-FedAvg, 
the clients train the input part of the model, and the server trains the 
output part of the model. In FedPer and FedRep, the clients train the output 
part of the model, and the server trains the input part of the model. In 
Ditto~\cite{Ditto} and pFedMe~\cite{NEURIPS2020_f4f1f13c}, the personalized 
models of the clients do not stray too far from the global model. 
Ditto updates the personalized models based on the difference between the model 
parameters of the global model and those of the personalized models, while 
pFedMe employs Moreau envelopes as the clients' regularized loss functions to 
facilitate convergence analysis.

\section{Experimental Design Dimensions}
\label{sec:ddesign_dimensinos}
In federated learning, datasets, client, and training settings affect the 
performance of learning methods. 
To evaluate the performance of existing methods and understand their 
characteristics, we consider the following three design dimensions in this 
study. 

\textbf{Number of clients.} 
The number of clients may significantly differ, depending on the use case 
we target. For example, the number of the clients may be around 10 for 
small institutions, while the number of the clients may be 100 or even more 
for mobile devices. As the number of clients increases, it becomes more 
difficult to aggregate models on the server, resulting in less accuracy. 
Therefore, a robust method for varying numbers of clients is desirable.

\textbf{Total number of data samples.} 
Like the number of clients, the total number of data samples also depends 
on the use case, and the performances of federated learning methods may 
differ when we vary the total number of data samples. Even if the server is 
aware of the numbers of data samples of the clients, it is challenging to 
select an optimal method. A robust method for different numbers of data 
samples is desirable. To this end, it is necessary to evaluate how the 
performances of existing methods vary with the total number of data samples.

\textbf{Degree of data heterogeneity.}
As the degree of data heterogeneity increases, the accuracy of 
non-personalized federated learning decreases, while personalized federated 
learning rather improves accuracy because it allows the construction of a 
model that fits each client. Previous studies have not comprehensively 
evaluated this impact on the performance of personalized federated learning 
methods. In this paper, we compare and discuss the accuracies of existing 
methods by varying the degree of data heterogeneity.

\section{Experiments}
\label{sec:3_experiments}
In this section, we introduce experimental configurations and report 
our experimental results. To answer the questions described in 
Section~\ref{sec:1_introduction}, we perform the following experiments: 
(1) To evaluate the performance of personalized and non-personalized 
federated learning methods, we compare the methods in terms of accuracy, 
convergence speed, communication traffic, and training time. 
(2) To evaluate the impact of experimental settings on accuracy, we 
conduct experiments by varying the number of clients, the total number 
of data samples, and the degree of data heterogeneity described in 
Section~\ref{sec:ddesign_dimensinos}.

To simplify the experiments, we used Pytorch to create a virtual client 
and the server on a single GPU machine. Experiments were performed on a 
Linux server with NVIDIA Tesla V100 SXM2 GPU (16GB) and Intel Xeon Gold 
6148 Processor CPU (384GB). 

\subsection{Experimental Setup}
\noindent
{\bf Datasets, tasks, and models.}
We use five datasets: FEMNIST, Shakespeare, Sent140, MNIST, and CIFAR-10, 
which are often used in previous studies~\cite{chen2021fedbe,FedRep,li2019fedmd,MLSYS2020_38af8613,mansour2020three,mcmahan2017communication,Wang2020Federated}.
FEMNIST, Shakespeare, and Sent140 are originally separated for federated 
learning. While since MNIST and CIFAR-10 are not separated, we need to 
divide these two datasets synthetically.

FEMNIST~\cite{caldas2018leaf} includes images of handwritten characters 
with $62$ labels, and is divided into 3,400 sub-datasets of writers. 
Shakespeare~\cite{MLSYS2020_38af8613} includes lines in ``The Complete 
Works of William Shakespeare'', and is divided into $143$ sub-datasets of 
actors. Sent140~\cite{caldas2018leaf} includes the text of tweets with 
$2$ labels, either positive sentiment or negative sentiment. This dataset 
is divided into 660,120 sub-datasets of twitter users, and we use $927$ 
sub-datasets with more than 50 tweets in the experiment. MNIST~\cite{MNIST} 
includes images of handwritten characters with $10$ labels. 
CIFAR-10~\cite{cifar} includes photo images with $10$ labels. We divide 
MNIST and CIFAR-10 into sub-datasets using the Dirichlet distribution as in 
\cite{Wang2020Federated}. Table~\ref{tab:statistics} shows the statistics 
of the above datasets. We note that we randomly divide MNIST and CIFAR-10 
in each test, so the statistics of them are the values in a single test.

\begin{table}[!t]
 \caption{Data statistics. Total size and test size indicate the numbers of 
 data samples in the entire dataset and the test data of the dataset, 
 respectively. Other measures are statistics of local datasets.}
 \label{tab:statistics}
 \centering
 \scalebox{1.0}{
 \begin{tabular}{ c c c c c c c} \Hline
Datasets & Total size & Test size & Mean & SD & Max & Min \\\hline
FEMNIST & 749{,}068 & 77{,}483 & 220.3 & 85.20 & 465 & 19 \\
Shakespeare & 517{,}106 & 103{,}477 & 3{,}616.1 & 6{,}832.37 & 41{,}305 & 3 \\
Sent140 & 74{,}589 & 7{,}895 & 80.5 & 40.02 & 549 & 50 \\
MNIST & 70{,}000 & 10{,}000 & 3{,}450.0 & 1{,}050.17 & 5{,}534 & 1{,}554 \\
CIFAR-10 & 60{,}000 & 10{,}000 & 2{,}950.0 & 1{,}233.60 & 6{,}043 & 1{,}360 \\
\Hline
\end{tabular}
}
\end{table}

In tasks and models, we follow the previous studies~\cite{chen2021fedbe,FedRep,li2019fedmd,MLSYS2020_38af8613,mansour2020three,mcmahan2017communication,reddi2020adaptive,Wang2020Federated}.
In task settings, we conduct an image classification task for FEMNIST, MNIST, and CIFAR-10.
For Shakespeare, we conduct a next-character prediction that infers the next characters after given sentences.
For Sent140, we conduct a binary classification that categorizes whether a tweet is a positive or negative sentiment.
We use different models for each task following the existing works~\cite{reddi2020adaptive,Wang2020Federated,FedRep}.
For FEMNIST and MNIST we use CNN, and for Shakespeare we use LSTM.
For CIFAR-10, we use VGG with the same modification reported in \cite{Wang2020Federated}.
For Sent140, we use a pre-trained 300-dimensional GloVe embedding\cite{Glove} and train RNN with an LSTM module.


\noindent
{\bf Client and training setting.}
We vary several parameters for clients and training: the number of clients, 
the total number of data samples, and the degree of data heterogeneity. 
The number of clients, $|S|$, is selected from $\{5,10,20,100\}$. We change 
the total number of data samples using a ratio $D$ to the entire dataset 
(i.e., the total number of data samples is $D \cdot N$), whose range is 
$\{0.25,0.5,0.75,0.1\}$. 
To change the degree of data heterogeneity, we use a parameter 
$\alpha_{label}$ to control the degree of heterogeneity for the labels on 
the clients. $\alpha_{label}$ is selected from $\{0.1,0.5,1.0,5.0\}$. 
The default values of $|S|$, $D$, and $\alpha_{label}$ are 20, 1.0, and 0.5, 
respectively. 


The five datasets are pre-partitioned into training and test data. 
In FEMNIST, Shakespeare, and Sent140, we randomly select $|S|$ 
sub-datasets as local data. In MNIST and CIFAR-10, we randomly divide 
the whole train and test data into $|S|$ local data. The distributions 
of test and train data follow the same Dirichlet distribution.
We split the training data into $7:3$ for FEMNIST, Shakespeare, and 
Sent140, and into $8:2$ for MNIST and CIFAR-10. The two splits are 
used for training and validation, respectively. We select 1,000 
unlabeled data from the training data for FedMe, and the unlabeled 
data is excluded from the training data.

We set the number of global communication rounds to be $300$, $200$, 
$500$, $100$, and $100$ for FEMNIST, MNIST, CIFAR-10, Shakespeare, and 
Sent140, respectively. We set the local epoch $E$ to be $2$ for all 
the settings. All the clients participate in each global communication 
round following recent studies~\cite{FedPer,shen2020federated,Wang2020Federated}.
We conduct training and test five times and report mean and standard 
deviation (std) of accuracy over five times of experiments with 
different clients.

\noindent
{\bf Methods and hyperparameter tuning.}
We compare three types of methods: (1) non-personalized federated 
learning methods, (2) personalized federated learning methods, and 
(3) non-federated learning methods. For (1), we use FedAvg and Fedprox; 
for (2), we use HypCluster, FML, FedMe, LG-FedAvg, FedPer, FedRep, 
Ditto, and pFedMe; for (3), we use Local Data Only, in which clients 
build their models on their local data, and Centralized, in which a 
server collects local data from all clients (centralized can be 
considered as an oracle). We use fine-tuning on each client for FedAvg, 
Fedprox, HypCluster, FedMe, and Centralized after building their models. 
In FML, LG-FedAvg, FedPer, FedRep, Ditto, and pFedMe, we do not use 
fine-tuning because techniques similar to fine-tuning are included in 
these methods. 

We describe hyperparameter tuning. The learning rate is selected from 
$\{10^{-3}, 10^{-2.5}, 10^{-2}, \ldots, 10^{0.5}\}$ and optimized for 
each method on default parameters. 
The optimized learning rate is used in the experiment of impact of the 
experimental setup. The optimization method is SGD (stochastic gradient 
descent) with momentum $0.9$ and weight decay $10^{-4}$. The batch 
sizes of FEMNIST, MNIST, CIFAR-10, Shakespeare, and Sent140 are $20$, 
$20$, $40$, $10$, and $4$, respectively. Hyperparameters specific to 
each method is described in the supplementary file.

\subsection{Performance Comparison}
We compare the methods in terms of accuracy, convergence speed, 
training speed, and communications traffic in the default parameter 
setting. In this experiment, we have the two findings: 

\begin{screen}
{\textbf{Finding 1.}  No method consistently outperforms the other methods in all the datasets.}
\end{screen}
\begin{screen}
{\textbf{Finding 2.}  Only a few state-of-the-art personalized methods outperform standard federated learning methods.}
\end{screen}
\noindent

{\bf Accuracy.}
Table~\ref{tab:result_accuracy} shows the accuracy and average ranking 
of each method. We note that the standard deviations of FEMNIST, Shakespeare, 
and Sent140 are relatively large because the clients differ in each 
test (we randomly select 20 clients from the set of clients). From 
Table~\ref{tab:result_accuracy}, we can see that the most accurate method is 
FedMe+FT for FEMNIST, Ditto for Shakespeare, Hypcluster for Sent140, 
FedAvg+FT for MNIST, and FedMe+FT for CIFAR-10. From this result, we find that 
none of the existing state-of-the-art personalized federated learning methods 
outperform the others in all the datasets.

We can also see that FedMe+FT has the highest average rank. On the other hand, 
the other personalized federated learning methods have lower average ranks 
than the standard federated learning methods such as FedAvg and FedProx with 
fine-tuning. From this result, we can find that only a few state-of-the-art 
personalized methods outperform standard federated learning methods, and those 
with fine-tuning are often sufficient to deal with data heterogeneity.

\begin{table}[!t]
\centering
\caption{Test accuracy (mean$\pm$std).}
\label{tab:result_accuracy}
\scalebox{0.89}{
\begin{tabular}{lccccc|c}\Hline
 & FEMNIST & Shakespeare & Sent140 & MNIST & CIFAR-10 & Average rank \\\hline
FedAvg & 75.79±1.65 & 44.94±1.96 & 58.83±11.88 & 98.90±0.10 & 86.05±0.48 & 9.2 \\
FedAvg+FT & 77.25±3.99 & 42.53±2.19 & 74.66±6.20 & \textbf{99.23±0.09} & 89.59±0.94 & 3.6 \\
FedProx & 76.08±2.12 & 48.59±3.59 & 58.83±11.88 & 98.87±0.06 & 86.01±0.38 & 8.6 \\
FedProx+FT & 76.96±3.42 & 45.17±2.83 & 74.66±6.20 & 99.20±0.10 & 89.76±0.62 & 3.6 \\\hdashline
HypCluster & 75.99±2.94 & 41.82±3.33 & \textbf{77.08±4.69 }& 98.90±0.09 & 85.21±1.22 & 7.4 \\
HypCluster+FT & 76.29±3.15 & 41.10±3.29 & 73.16±9.41 & 99.15±0.12 & 88.54±1.42 & 7.2 \\
FML & 67.91±2.53 & 28.73±1.78 & 72.49±8.87 & 98.26±0.16 & 79.89±1.44 & 12.0 \\
FedMe & 77.64±2.39 & 46.98±2.30 & 73.99±8.29 & 98.92±0.14 & 88.15±0.52 & 5.8 \\
FedMe+FT & \textbf{78.06±3.00} & 45.83±2.48 & 74.41±8.16 & 99.17±0.07 & \textbf{90.96±0.84} & \textbf{2.8} \\
LG-FedAvg & 65.14±3.12 & 23.17±1.93 & 73.41±10.07 & 97.80±0.16 & 78.53±1.57 & 13.0 \\
FedPer & 65.96±2.81 & 30.83±3.32 & 74.16±7.59 & 99.11±0.08 & 90.00±0.83 & 8.0 \\
FedRep & 66.04±2.20 & 31.71±2.29 & 73.91±8.33 & 99.06±0.07 & 88.96±0.48 & 8.8 \\
Ditto & 75.68±3.63 & \textbf{49.33±1.85} & 74.28±8.10 & 99.22±0.06 & 90.41±0.67 & 3.8 \\
pFedMe & 72.92±3.54 & 40.33±2.27 & 71.20±10.25 & 98.96±0.05 & 79.46±2.08 & 10.6 \\\hdashline
Local Data Only & 64.71±2.94 & 24.77±1.95 & 74.33±7.86 & 97.60±0.28 & 73.17±1.55 & - \\
Centralized & 76.08±1.65 & 47.64±2.63 & 58.83±11.88 & 98.89±0.05 & 85.96±0.54 & - \\
Centralized+FT & 79.35±2.29 & 48.43±3.32 & 67.91±7.41 & 99.27±0.08 & 90.80±0.92 & - \\
\Hline
\end{tabular}
}
\end{table}

\begin{figure}[!t]
 \centering
    \begin{minipage}[t]{1.0\linewidth}
        \centering
        \includegraphics[width=1.0\linewidth]{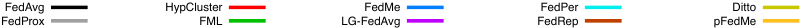}
    \end{minipage}
    \\
    \begin{minipage}[t]{0.30\linewidth}
        \centering
        \includegraphics[width=1.0\linewidth]{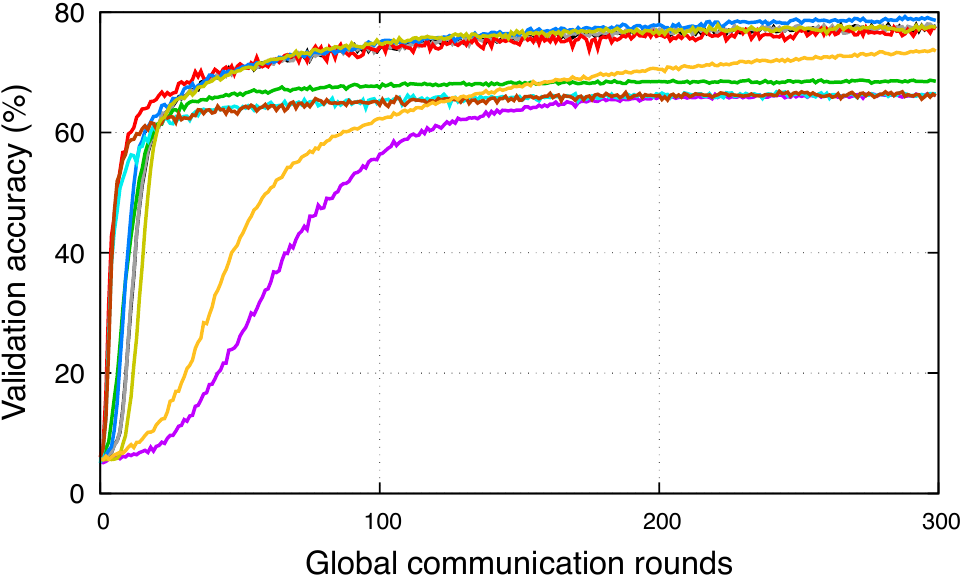}
        \subcaption{FEMNIST}
        \label{fig:femnist_result}
    \end{minipage}
    \begin{minipage}[t]{0.30\linewidth}
        \centering
        \includegraphics[width=1.0\linewidth]{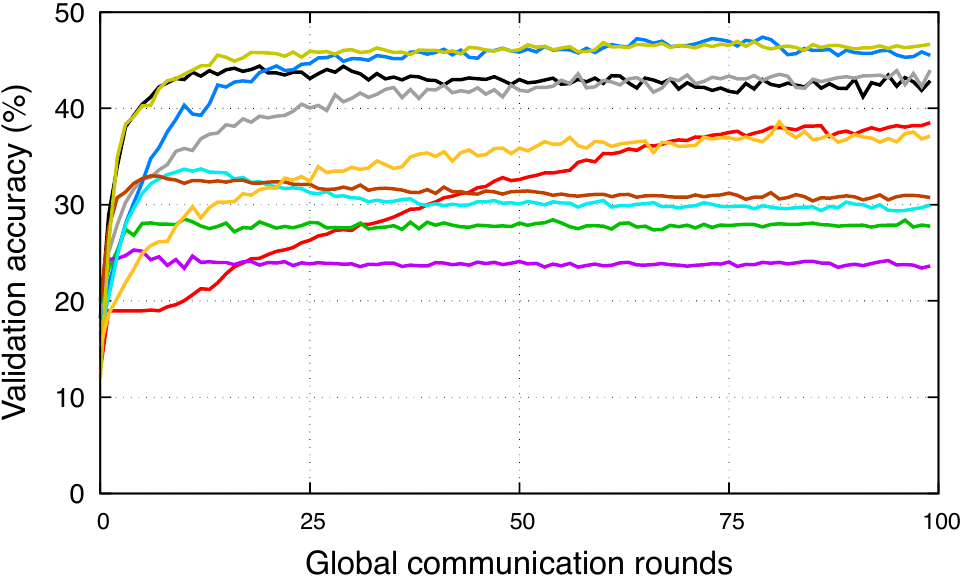}
        \subcaption{Shakespeare}
        \label{fig:shakespeare_result}
    \end{minipage}
    \begin{minipage}[t]{0.30\linewidth}
        \centering
        \includegraphics[width=1.0\linewidth]{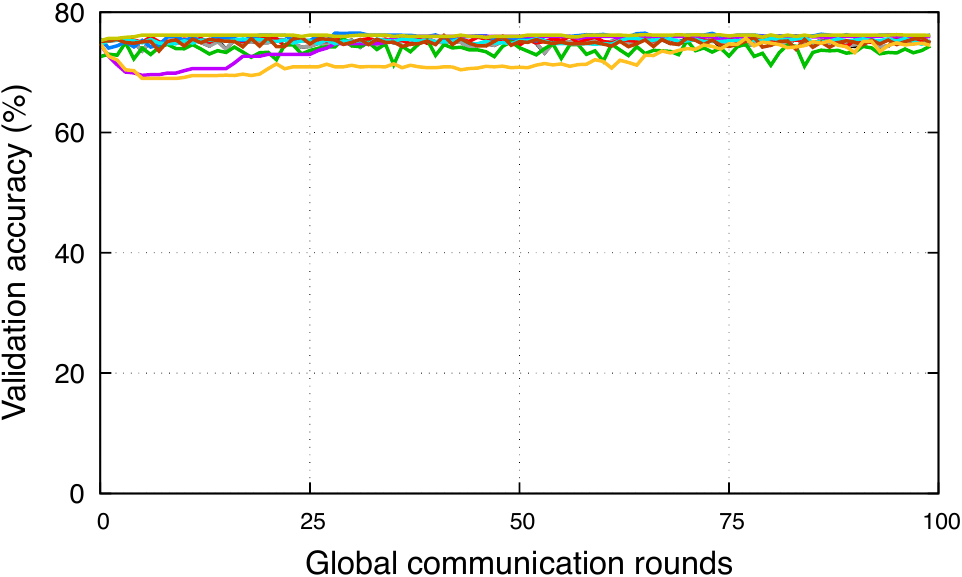}
        \subcaption{Sent140}
        \label{fig:sent140_result}
    \end{minipage} \\
    \vspace{3 mm}
    \begin{minipage}[t]{0.30\linewidth}
        \centering
        \includegraphics[width=1.0\linewidth]{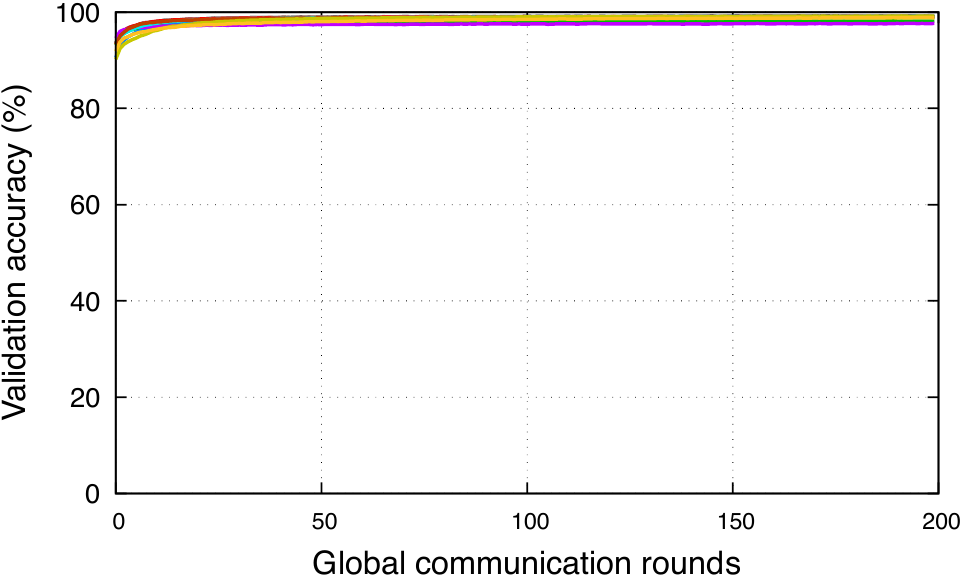}
        \subcaption{MNIST}
        \label{fig:mnist_result}
    \end{minipage}
    \begin{minipage}[t]{0.30\linewidth}
        \centering
        \includegraphics[width=1.0\linewidth]{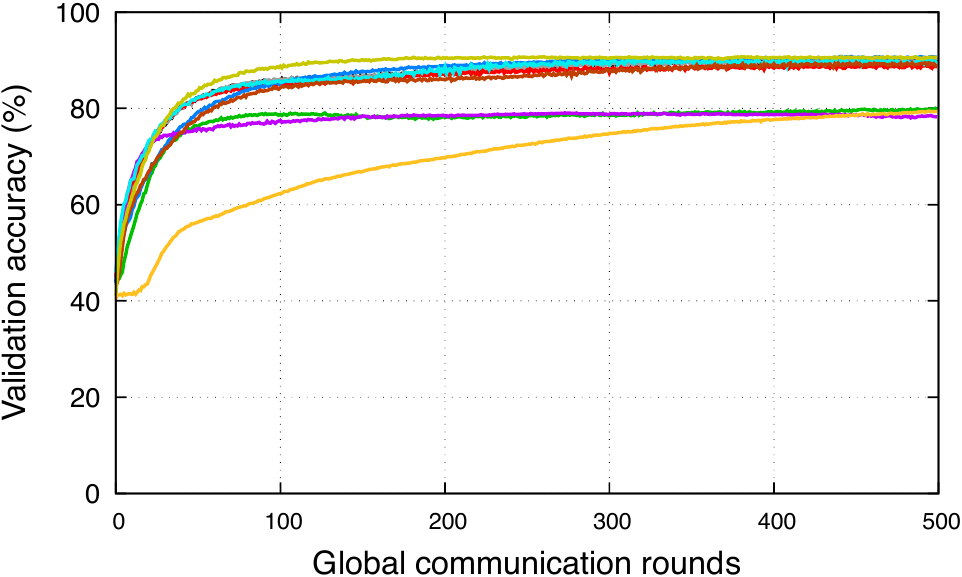}
        \subcaption{CIFAR-10}
        \label{fig:cifar10_result}
    \end{minipage}
\caption{The validation accuracy over time of various methods.}
\label{fig:convergence_speed}
\end{figure}

\noindent
{\bf Convergence speed.}
Figure~\ref{fig:convergence_speed} shows the validation accuracy 
of each global communication round. The validation accuracy is the 
average accuracy at each epoch of the five experiments. Since each 
client evaluates its model by its own validation data after training 
its model and before aggregating models, the accuracy of each method 
is equivalent to that after fine-tuning.


From Figure~\ref{fig:convergence_speed}, we can see that FedAvg and 
Ditto are stable and converge quickly for all datasets. On the other 
hand, we can see that FedMe has the highest average rank but loses 
in convergence speed to FedAvg and Ditto. From this result, we can 
find that the methods with the highest accuracy and the fastest 
convergence are different.

\begin{figure}[!t]
 \centering
    \begin{minipage}[t]{1.0\linewidth}
        \centering
        \includegraphics[width=1.0\linewidth]{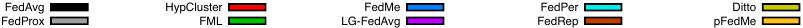}
    \end{minipage}
    \\
    \begin{minipage}[t]{1.0\linewidth}
        \centering
        \includegraphics[width=0.9\linewidth]{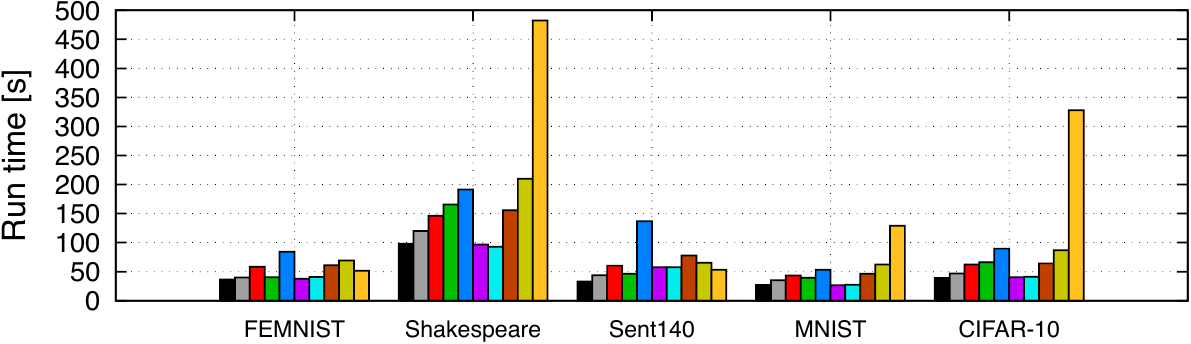}
    \end{minipage}
\caption{Training time per global communication round.}
\label{fig:training_speed}
\end{figure}

\noindent
{\bf Training time.}
We evaluate run time on the training phase in each method. 
Figure~\ref{fig:training_speed} shows the average run time per 
global communication round. We note that the run time is the 
average of ten global communication rounds.


From Figure~\ref{fig:training_speed}, we can see that FedAvg 
has the smallest training time for all datasets. FedMe and Ditto 
have a large training time than the other methods. pFedMe spends 
similar training time to the other methods on FEMNIST and 
Sent140, while it spends much larger time than the other methods 
on Shakespeare, MNIST, and CIFAR-10. pFedMe has large training 
time for clients, so when the volume of local data increases, 
its training time increases.


\noindent
{\bf Communications traffic.}
We evaluate communications traffic on the training phase in each method. 
Since each method exchanges models between the server and client, 
communications traffic is compared by the size of model parameters sent 
per global communication round. Figure~\ref{tab:communications_traffic} 
shows the communications traffic per global communication round. 


From Figure~\ref{tab:communications_traffic}, we can see that FedMe has 
the largest communication traffic. This is because FedMe has the extra 
model transmission compared with the other methods. FedPer, FedRep, and 
LG-FedAvg have smaller communication traffic than other methods because 
these three methods send only a part of the model between the server and 
the clients. LG-FedAvg has the smallest communication traffic among them 
because the output side of the model has a smaller number of model 
parameters than the input side of the model.

\begin{table}[!t]
\centering
\caption{Communication traffic: the number of model parameters communicated between the server and the clients per round.}
\label{tab:communications_traffic}
\scalebox{0.69}{
\begin{tabular}{lclclclclcl}\Hline
 & FEMNIST &  & Shakespeare &  & Sent140 &  & MNIST &  & CIFAR-10 &  \\\hline
FedAvg & 2413180 &  (1$\times$) & 1645140 &  (1$\times$) & 161344 &  (1$\times$) & 2399764 &  (1$\times$) & 19870868 &  (1$\times$) \\
FedProx & 2413180 &  (1$\times$) & 1645140 &  (1$\times$) & 161344 &  (1$\times$) & 2399764 &  (1$\times$) & 19870868 &  (1$\times$) \\\hdashline
HypCluster & 3619770 &  (1.5$\times$) & 2467710 &  (1.5$\times$) & 242016 &  (1.5$\times$) & 3599646 &  (1.5$\times$) & 29806302 &  (1.5$\times$) \\
FML & 2413180 &  (1$\times$) & 1645140 &  (1$\times$) & 161344 &  (1$\times$) & 2399764 &  (1$\times$) & 19870868 &  (1$\times$) \\
FedMe & 6032950 &  (2.5$\times$) & 4112850 &  (2.5$\times$) & 403360 &  (2.5$\times$) & 5999410 &  (2.5$\times$) & 49677170 &  (2.5$\times$) \\
LG-FedAvg & 15996 &  (0.007$\times$) & 46260 &  (0.028$\times$) & 25644 &  (0.159$\times$) & 2580 &  (0.001$\times$) & 1060884 &  (0.053$\times$) \\
FedPer & 2397184 &  (0.993$\times$) & 1598880 &  (0.972$\times$) & 161300 &  (1$\times$) & 2397184 &  (0.999$\times$) & 18809984 &  (0.947$\times$) \\
FedRep & 2397184 &  (0.993$\times$) & 1598880 &  (0.972$\times$) & 161300 &  (1$\times$) & 2397184 &  (0.999$\times$) & 18809984 &  (0.947$\times$) \\
Ditto & 2413180 &  (1$\times$) & 1645140 &  (1$\times$) & 161344 &  (1$\times$) & 2399764 &  (1$\times$) & 19870868 &  (1$\times$) \\
pFedMe & 2413180 &  (1$\times$) & 1645140 &  (1$\times$) & 161344 &  (1$\times$) & 2399764 &  (1$\times$) & 19870868 &  (1$\times$) \\
\Hline
\end{tabular}
}
\end{table}


\subsection{Impact of Experimental Settings on Accuracy}
In this section, we compare the accuracy of each method in different experimental settings.

\noindent
{\bf Impact of the number of clients.}
Table~\ref{tab:impact_number_clients} show the accuracy of varying the number of clients.


From Table~\ref{tab:impact_number_clients}, we can see that the accuracy decreases significantly as the number of clients increases.
As the number of clients increases, it becomes more difficult to aggregate the model on the server, resulting in decreasing accuracy. 
FedAvg+FT has the highest average rank for MNIST, and Ditto has the highest average rank for CIFAR-10.
This result indicates that the larger number of clients is more challenging, while we can design robust methods for different number of clients.


\begin{table}[!t]
\centering
\caption{Accuracy v.s. number of clients.}
\label{tab:impact_number_clients}
\scalebox{0.55}{
\begin{tabular}{lccccccccccc}\Hline
 & \multicolumn{5}{c}{MNIST} & & \multicolumn{5}{c}{CIFAR-10}\\
 \cline{2-6} \cline{8-12}
 & |S|=5 & |S|=10 & |S|=20 & |S|=100 & Average rank &  & |S|=5 & |S|=10 & |S|=20 & |S|=100 & Average rank \\\hline
FedAvg & 98.68±0.13 & 98.82±0.14 & 98.90±0.10 & 98.67±0.08 & 9.3 &  & 87.33±0.56 & 86.62±1.11 & 86.05±0.48 & 81.60±0.50 & 8.8 \\
FedAvg+FT & 99.20±0.16 & \textbf{99.26±0.06} & \textbf{99.23±0.09} & 98.87±0.09 & \textbf{1.8} &  & 90.18±1.05 & 90.89±0.51 & 89.59±0.94 & 82.61±0.58 & 4.0 \\
FedProx & 98.65±0.11 & 98.81±0.14 & 98.87±0.06 & 98.66±0.09 & 10.5 &  & 87.82±0.94 & 86.80±0.70 & 86.01±0.38 & 81.55±0.53 & 8.8 \\
FedProx+FT & 99.17±0.14 & 99.24±0.11 & 99.20±0.10 & \textbf{98.89±0.12} & 2.8 &  & 90.22±1.04 & 91.16±0.54 & 89.76±0.62 & 82.24±0.83 & 3.5 \\\hdashline
HypCluster & 98.86±0.20 & 98.86±0.08 & 98.90±0.09 & 98.62±0.08 & 9.0 &  & 86.27±0.79 & 85.71±1.22 & 85.21±1.22 & 80.38±1.01 & 10.8 \\
HypCluster+FT & 99.12±0.15 & 99.11±0.11 & 99.15±0.12 & 98.84±0.14 & 5.5 &  & 88.52±1.00 & 88.60±0.92 & 88.54±1.42 & 81.65±0.29 & 6.8 \\
FML & 98.84±0.31 & 98.71±0.13 & 98.26±0.16 & 95.31±0.35 & 12.5 &  & 84.01±7.31 & 83.80±0.56 & 79.89±1.44 & 65.97±0.68 & 12.3 \\
FedMe & 98.93±0.14 & 98.87±0.12 & 98.92±0.14 & 97.93±0.20 & 9.3 &  & 89.68±0.65 & 89.15±1.11 & 88.15±0.52 & 76.55±1.21 & 8.0 \\
FedMe+FT & 99.23±0.15 & 99.19±0.12 & 99.17±0.07 & 98.35±0.19 & 4.5 &  & 91.81±0.85 & \textbf{91.93±0.24} & \textbf{90.96±0.84} & 81.23±0.47 & 3.0 \\
LG-FedAvg & 87.10±4.49 & 96.46±3.97 & 97.80±0.16 & 94.30±0.41 & 14.0 &  & 85.47±2.12 & 83.01±1.01 & 78.53±1.57 & 64.85±1.00 & 13.0 \\
FedPer & 99.15±0.20 & 99.21±0.08 & 99.11±0.08 & 97.96±0.26 & 6.0 &  & 90.27±1.81 & 90.33±0.85 & 90.00±0.83 & 81.61±0.80 & 4.0 \\
FedRep & 99.13±0.26 & 99.11±0.07 & 99.06±0.07 & 97.86±0.27 & 8.0 &  & 89.57±1.58 & 89.52±0.60 & 88.96±0.48 & 80.51±0.69 & 7.0 \\
Ditto & \textbf{99.27±0.17} & 99.25±0.08 & 99.22±0.06 & 98.19±0.24 & 3.3 &  & \textbf{92.15±0.76} & 91.59±0.40 & 90.41±0.67 & \textbf{82.79±0.68} & \textbf{1.5} \\
pFedMe & 99.14±0.17 & 99.08±0.09 & 98.96±0.05 & 97.89±0.19 & 8.3 &  & 81.08±2.73 & 82.13±2.53 & 79.46±2.08 & 58.41±1.63 & 13.8 \\\hdashline
Local Data Only & 98.78±0.27 & 98.28±0.12 & 97.60±0.28 & 93.86±0.37 & - &  & 82.23±2.26 & 78.31±1.26 & 73.17±1.55 & 59.76±1.03 & - \\
Centralized & 98.83±0.06 & 98.91±0.06 & 98.89±0.05 & 98.81±0.06 & - &  & 85.93±0.28 & 86.24±0.60 & 85.96±0.54 & 86.07±0.55 & - \\
Centralized+FT & 99.21±0.10 & 99.24±0.03 & 99.27±0.08 & 99.06±0.06 & - &  & 90.16±1.13 & 90.91±0.36 & 90.80±0.92 & 90.62±0.40 & - \\

\Hline
\end{tabular}
}
\end{table}


\noindent
{\bf Impact of the total number of data samples.}
Table~\ref{tab:impact_total number_data} shows the accuracy when 
we vary the total number of data samples.


From Table~\ref{tab:impact_total number_data}, we can see that the 
accuracy decreases as the total number of data samples decreases. 
This is because clients do not have sufficient data samples to train 
their models when the number of data samples is small. 
The ranks of methods do not change much, so the number of data 
samples does not significantly impact deciding the superiority of 
methods.

\begin{table}[!t]
\centering
\caption{Accuracy v.s. total number of data samples.}
\label{tab:impact_total number_data}
\scalebox{0.55}{
\begin{tabular}{lccccccccccc}\Hline
 & \multicolumn{5}{c}{MNIST} & & \multicolumn{5}{c}{CIFAR-10}\\
 \cline{2-6} \cline{8-12}
 & D=0.25 & D=0.5 & D=0.75 & D=1.0 & Average rank &  & D=0.25 & D=0.5 & D=0.75 & D=1.0 & Average rank \\\hline
FedAvg & 97.76±0.30 & 98.49±0.26 & 98.70±0.07 & 98.90±0.10 & 8.5 &  & 70.82±1.58 & 79.99±0.41 & 83.75±0.71 & 86.05±0.48 & 9.5 \\
FedAvg+FT & \textbf{98.41±0.19} & 98.90±0.24 & \textbf{99.13±0.06} & \textbf{99.23±0.09} & \textbf{1.3} &  & 77.27±2.19 & 84.76±0.23 & 87.75±1.09 & 89.59±0.94 & 4.0 \\
FedProx & 97.79±0.30 & 98.56±0.24 & 98.68±0.08 & 98.87±0.06 & 8.8 &  & 71.13±1.35 & 79.90±0.42 & 83.84±0.73 & 86.01±0.38 & 9.5 \\
FedProx+FT & 98.36±0.23 & \textbf{98.98±0.17} & \textbf{99.13±0.05} & 99.20±0.10 & 1.8 &  & 76.75±1.71 & 84.61±0.47 & 88.17±1.08 & 89.76±0.62 & 4.0 \\\hdashline
HypCluster & 97.20±0.20 & 98.44±0.23 & 98.66±0.15 & 98.90±0.09 & 10.5 &  & 70.38±0.84 & 78.83±0.61 & 82.59±1.39 & 85.21±1.22 & 11.0 \\
HypCluster+FT & 97.73±0.21 & 98.66±0.28 & 98.94±0.13 & 99.15±0.12 & 6.0 &  & 73.84±2.30 & 82.28±0.75 & 85.95±1.35 & 88.54±1.42 & 7.3 \\
FML & 95.53±1.94 & 97.23±0.30 & 97.90±0.12 & 98.26±0.16 & 13.0 &  & 65.86±3.54 & 72.99±0.86 & 77.38±1.88 & 79.89±1.44 & 12.3 \\
FedMe & 97.83±0.38 & 98.47±0.13 & 98.63±0.13 & 98.92±0.14 & 9.0 &  & 71.43±3.49 & 83.50±1.15 & 86.45±1.10 & 88.15±0.52 & 7.3 \\
FedMe+FT & 98.11±0.45 & 98.78±0.19 & 99.01±0.08 & 99.17±0.07 & 3.8 &  & \textbf{78.21±1.32} & \textbf{86.40±0.63} & \textbf{89.43±0.93} & \textbf{90.96±0.84} & \textbf{1.0} \\
LG-FedAvg & 94.58±2.11 & 96.54±0.25 & 97.35±0.18 & 97.80±0.16 & 14.0 &  & 66.20±3.12 & 72.26±0.97 & 76.38±2.22 & 78.53±1.57 & 13.3 \\
FedPer & 97.00±1.69 & 98.62±0.25 & 98.99±0.10 & 99.11±0.08 & 7.3 &  & 77.39±2.53 & 85.26±0.86 & 88.32±0.47 & 90.00±0.83 & 2.3 \\
FedRep & 97.10±1.65 & 98.60±0.09 & 98.94±0.08 & 99.06±0.07 & 7.8 &  & 75.65±1.98 & 83.82±0.72 & 87.36±0.96 & 88.96±0.48 & 5.5 \\
Ditto & 98.01±0.35 & 98.86±0.20 & 99.09±0.08 & 99.22±0.06 & 3.0 &  & 73.41±2.72 & 83.40±1.25 & 88.18±0.71 & 90.41±0.67 & 4.8 \\
pFedMe & 97.58±0.23 & 98.39±0.27 & 98.67±0.14 & 98.96±0.05 & 9.8 &  & 60.95±3.73 & 71.85±1.05 & 77.00±1.66 & 79.46±2.08 & 13.5 \\\hdashline
Local Data Only & 94.17±1.98 & 96.07±0.26 & 97.05±0.14 & 97.60±0.28 & - &  & 60.34±4.30 & 66.50±1.01 & 70.51±2.61 & 73.17±1.55 & - \\
Centralized & 97.72±0.28 & 98.56±0.19 & 98.77±0.10 & 98.89±0.05 & - &  & 74.67±1.40 & 80.53±0.75 & 83.97±0.98 & 85.96±0.54 & - \\
Centralized+FT & 98.27±0.13 & 98.95±0.20 & 99.12±0.05 & 99.27±0.08 & - &  & 83.21±0.76 & 87.06±0.58 & 89.64±1.03 & 90.80±0.92 & - \\

\Hline
\end{tabular}
}
\end{table}


\noindent
{\bf Impact of the degree of data heterogeneity.}
Table~\ref{tab:impact_data_heterogeneity} shows the accuracy when 
we vary the degree of data heterogeneity. A smaller $\alpha_{label}$ 
indicates a larger degree of data heterogeneity.

\begin{screen}
{\textbf{Finding 3. The larger the degree of data heterogeneity, the 
more accurate the personalized federated learning methods are.} }
\end{screen}

From Table~\ref{tab:impact_data_heterogeneity}, we can see that the 
accuracy of FedAvg and FedProx decreases as the degree of data 
heterogeneity increases. On the other hand, we can see that the 
accuracy of personalized federated learning methods tends to 
increase as the degree of data heterogeneity increases. As the degree 
of data heterogeneity increases, the clients can easily build their 
personalized models that fit their local data. We can find that data 
heterogeneity works positively for personalized federated learning.

We can also see that FedAvg+FT and FedProx+FT have the highest average 
rank on MNIST, and FedMe+FT has the highest average rank on CIFAR-10. 
This result indicates that the standard federated learning methods 
with fine-tuning are often sufficient to deal with the data heterogeneity.

\begin{table}[!t]
\centering
\caption{Accuracy v.s. degree of data heterogeneity.}
\label{tab:impact_data_heterogeneity}
\scalebox{0.55}{
\begin{tabular}{lccccccccccc}\Hline
 & \multicolumn{5}{c}{MNIST} & & \multicolumn{5}{c}{CIFAR-10}\\
 \cline{2-6} \cline{8-12}
 & $\alpha_{label}$=5.0 & $\alpha_{label}$=1.0 & $\alpha_{label}$=0.5 & $\alpha_{label}$=0.1 & Average rank &  & $\alpha_{label}$=5.0 & $\alpha_{label}$=1.0 & $\alpha_{label}$=0.5 & $\alpha_{label}$=0.1 & Average rank \\\hline
FedAvg & 98.95±0.04 & 98.89±0.09 & 98.90±0.10 & 98.61±0.24 & 7.0 &  & 86.76±0.40 & 86.20±0.66 & 86.05±0.48 & 80.04±2.89 & 9.5 \\
FedAvg+FT & 98.94±0.08 & \textbf{99.12±0.06} & \textbf{99.23±0.09} & 99.52±0.18 & \textbf{1.8} &  & 86.67±0.75 & 88.56±1.32 & 89.59±0.94 & 94.45±0.97 & 4.5 \\
FedProx & 98.93±0.05 & 98.90±0.06 & 98.87±0.06 & 98.61±0.23 & 7.8 &  & 86.79±0.21 & 86.26±0.39 & 86.01±0.38 & 80.86±2.02 & 9.0 \\
FedProx+FT & \textbf{98.98±0.06} & 99.07±0.10 & 99.20±0.10 & \textbf{99.54±0.14} & \textbf{1.8} &  & 86.17±0.46 & 88.51±0.65 & 89.76±0.62 & 94.57±1.11 & 4.5 \\\hdashline
HypCluster & 98.87±0.16 & 98.50±0.58 & 98.90±0.09 & 98.38±0.26 & 9.5 &  & 84.93±0.45 & 84.45±0.64 & 85.21±1.22 & 82.14±1.88 & 11.0 \\
HypCluster+FT & 98.81±0.09 & 98.71±0.64 & 99.15±0.12 & 99.40±0.12 & 6.3 &  & 84.13±0.49 & 86.52±1.08 & 88.54±1.42 & 93.92±1.20 & 8.0 \\
FML & 97.79±0.16 & 96.92±1.79 & 98.26±0.16 & 98.01±1.83 & 13.0 &  & 68.89±0.89 & 75.59±1.50 & 79.89±1.44 & 91.16±2.29 & 11.3 \\
FedMe & 98.72±0.11 & 98.73±0.20 & 98.92±0.14 & 98.89±0.25 & 8.0 &  & 87.01±0.45 & 87.98±0.68 & 88.15±0.52 & 82.79±8.33 & 7.0 \\
FedMe+FT & 98.84±0.09 & 98.93±0.24 & 99.17±0.07 & 99.46±0.12 & 4.5 &  & \textbf{87.73±0.45} & \textbf{89.60±0.74} & \textbf{90.96±0.84} & 94.50±1.28 & \textbf{1.5} \\
LG-FedAvg & 97.08±0.11 & 96.28±1.61 & 97.80±0.16 & 97.77±1.94 & 14.0 &  & 67.66±0.76 & 74.26±1.63 & 78.53±1.57 & 90.93±2.28 & 12.5 \\
FedPer & 98.81±0.07 & 97.80±1.71 & 99.11±0.08 & 98.84±0.98 & 8.5 &  & 86.25±0.79 & 88.26±0.73 & 90.00±0.83 & 93.97±1.70 & 5.0 \\
FedRep & 98.71±0.08 & 97.78±1.82 & 99.06±0.07 & 98.86±1.05 & 9.5 &  & 85.25±0.55 & 86.81±0.98 & 88.96±0.48 & 93.55±1.69 & 7.3 \\
Ditto & 98.89±0.08 & 99.05±0.15 & 99.22±0.06 & 98.87±1.44 & 4.3 &  & 87.52±0.34 & 89.22±0.32 & 90.41±0.67 & \textbf{94.82±1.06} & 1.8 \\
pFedMe & 98.67±0.09 & 98.69±0.19 & 98.96±0.05 & 99.21±0.09 & 8.5 &  & 69.93±1.08 & 63.16±29.82 & 79.46±2.08 & 86.23±3.48 & 12.3 \\\hdashline
Local Data Only & 96.73±0.14 & 96.02±1.58 & 97.60±0.28 & 97.76±1.54 & - &  & 59.03±0.34 & 66.74±1.32 & 73.17±1.55 & 88.85±2.92 & - \\
Centralized & 98.85±0.06 & 98.90±0.02 & 98.89±0.05 & 98.83±0.13 & - &  & 85.68±0.70 & 85.62±0.75 & 85.96±0.54 & 85.84±0.74 & - \\
Centralized+FT & 99.02±0.08 & 99.11±0.10 & 99.27±0.08 & 99.37±0.16 & - &  & 87.27±0.31 & 88.99±0.67 & 90.80±0.92 & 95.59±1.10 & - \\
\Hline
\end{tabular}
}
\end{table}


\subsection{Summary of Experimental Results}
We summarize the results of the above experimental study. First, 
there is a trade-off between accuracy, communication traffic, and 
training time. For example, FedMe is accurate in various 
experimental settings but reports large communication traffic and 
training time. Second, the standard federated learning methods with 
fine-tuning can deal well with data heterogeneity. In particular, 
for easy-to-learn datasets such as MNIST, they outperform the 
personalized federated learning methods. Finally, for a small 
number of clients, a large total number of data samples, or a 
large degree of heterogeneity, we observed higher accuracies of 
federated learning methods. These characteristics should be 
considered when developing and evaluating new federated learning 
methods.

\section{Conclusions, Limitations, and Future Work}
\label{sec:4_conclusion}
We evaluated personalized federated learning in various experimental settings. 
The experimental results showed several key findings: First, no method 
consistently outperformed the others in all the datasets. Second, the large 
degree of data heterogeneity improved the accuracy of personalized federated 
learning methods. Third, standard federated learning with fine-tuning was 
accurate compared with most personalized federated learning methods. We opened 
our Jupyter notebook-based tool {\sf FedBench} to facilitate experimental 
studies. 

This study has three limitations. First, despite 17 methods (ten federated 
learning, four variants, and three non-federated learning methods) and five 
datasets were used in this study, which are comprehensive compared with 
previous ones, we also note that there are numerous other federated 
learning methods (e.g., ~\cite{pmlr-v139-acar21a,APFL,fallah2020personalized,l2gd,huang2021personalized,karimireddy2020scaffold,marfoq2021federated,ozkara2021quped,smith2017federated,fednova,fedfomo}) and 
datasets (e.g., DigitFive and Office-Caltech10~\cite{li2021fedbn}, 
PROSTATE~\cite{liu2020shape}, Flicker mammal~\cite{flicker-mammal}, and 
FLCKER-AES and REAL-CUR~\cite{ren2017personalized}). Second, to study 
the impact of the data heterogeneity, we controlled the label distribution 
skew but did not investigate the impact of other types of skews, such 
as quantity skew, in which each client has a different number of data samples, 
and feature distribution skew, in which the clients' data share the same 
labels but vary in features. Third, we varied the number of clients, the 
total number of data samples, and the degree of data heterogeneity, whereas 
other parameters, such as client participant ratio, the number of local 
epochs, and model architectures, were not varied.

As future work, we plan to enrich our benchmark tool by addressing the above 
limitations and find further insights. We hope that our benchmark tool and 
experimental results help to develop and evaluate new federated learning methods.

\section*{Acknowledgement}
\label{sec:5_acknowledgement}
This work was supported by JST PRESTO Grant Number JPMJPR21C5 and JSPS KAKENHI Grant Number JP20H00584 and JP17H06099, Japan.

{
\small
\bibliographystyle{abbrv}
\vspace{-3mm}
\bibliography{bibliography}
}

\appendix


\section{Dataset statistics}
\label{appendix:data_statistics}
Detailed statistics for the datasets used in our experiments are shown in Table~\ref{tab:data_stats}.

\begin{table}[h]
\centering
\caption{Data Statistics for train and test data.}
\label{tab:data_stats}
\scalebox{0.65}{
\begin{tabular}{lcccccccccccccc}\Hline
 & \multicolumn{2}{c}{Size} & & \multicolumn{2}{c}{Mean} & & \multicolumn{2}{c}{SD} & & \multicolumn{2}{c}{Max} & & \multicolumn{2}{c}{Min}\\
 \cline{2-3}  \cline{5-6}  \cline{8-9} \cline{11-12} \cline{14-15}
 & Train & Test &  & Train & Test &  & Train & Test &  & Train & Test &  & Train & Test \\\hline
FEMNIST & 671{,}585 & 77{,}483 &  & 197.53 & 22.79 &  & 76.69 & 8.51 &  & 418 & 47 &  & 16 & 3 \\
Shakespeare & 413{,}629 & 103{,}477 &  & 2{,}892.51 & 723.62 &  & 5{,}465.89 & 1{,}366.48 &  & 33{,}044 & 8{,}261 &  & 2 & 1 \\
Sent140 & 66{,}694 & 7{,}895 &  & 71.95 & 8.52 &  & 36.04 & 4.00 &  & 494 & 55 &  & 45 & 5 \\
MNIST & 59{,}000 & 10{,}000 &  & 2950.00 & 500.00 &  & 896.15 & 154.09 &  & 4725 & 809 &  & 1{,}334 & 220 \\
CIFAR-10 & 49{,}000 & 10{,}000 &  & 2450.00 & 500.00 &  & 1{,}024.66 & 208.95 &  & 5{,}018 & 1{,}025 &  & 1{,}131 & 229 \\
\Hline
\end{tabular}
}
\end{table}

\section{Model architectures}
\label{appendix:model_architecture}
The details of the model used in our experiments are shown in Tables~\ref{tab:model_femnist}--\ref{tab:model_cifar}.

\begin{table}[h]
\centering
\caption{Model architectures for FEMNIST.}
\label{tab:model_femnist}
\scalebox{0.9}{
\begin{tabular}{lcccc}\Hline
Layer & Output Shape & Trainable Parameters & Activation & Hyperparameters \\\hline
Input & (28, 28, 1) & 0 & - & - \\
Conv2d & (26, 26, 32) & 320 & - & kernel size = 3; strides=(1, 1) \\
Conv2d & (24, 24, 64) & 18496 & - & kernel size = 3; strides=(1, 1) \\
MaxPool2d & (12, 12, 64) & 0 & - & pool size= (2, 2) \\
Dropout & (12, 12, 64) & 0 & - & p = 0.25 \\
Flatten & 9216 & 0 & - & - \\
Linear & 128 & 1179776 & ReLU & - \\
Dropout & 128 & 0 & - & p = 0.5 \\
Linear & 62 & 7998 & Softmax & - \\
\Hline
\end{tabular}
}
\end{table}

\begin{table}[h]
\centering
\caption{Model architectures for Shakespeare.}
\label{tab:model_shake}
\scalebox{0.9}{
\begin{tabular}{lcccc}\Hline
Layer & Output Shape & Trainable Parameters & Activation & Hyperparameters \\\hline
Input & 80 & 0 & - & - \\
Embedding & (80, 8) & 720 & - & - \\
LSTM & (80, 256) & 798720 & - & number of layers = 2 \\
Linear & 90 & 23130 & Softmax & - \\
\Hline
\end{tabular}
}
\end{table}

\begin{table}[h]
\centering
\caption{Model architectures for Sent140.}
\label{tab:model_sent}
\scalebox{0.9}{
\begin{tabular}{lcccc}\Hline
Layer & Output Shape & Trainable Parameters & Activation & Hyperparameters \\\hline
Input & 25 & 0 & - & - \\
Embedding & (300, 25) & 0 & - & - \\
LSTM & (300, 128) & 79360 & - & number of layers = 1 \\
Linear & (300, 10) & 1290 & ReLU & - \\
Dropout & (300, 10) & 0 & - & p = 0.5 \\
Linear & 2 & 22 & Softmax & - \\
\Hline
\end{tabular}
}
\end{table}

\begin{table}[h]
\centering
\caption{Model architectures for MNIST.}
\label{tab:model_mnist}
\scalebox{0.9}{
\begin{tabular}{lcccc}\Hline
Layer & Output Shape & Trainable Parameters & Activation & Hyperparameters \\\hline
Input & (28, 28, 1) & 0 & - & - \\
Conv2d & (26, 26, 32) & 320 & - & kernel size = 3; strides=(1, 1) \\
Conv2d & (24, 24, 64) & 18{,}496 & - & kernel size = 3; strides=(1, 1) \\
MaxPool2d & (12, 12, 64) & 0 & - & pool size= (2, 2) \\
Dropout & (12, 12, 64) & 0 & - & p = 0.25 \\
Flatten & 9{,}216 & 0 & - & - \\
Linear & 128 & 1{,}179{,}776 & ReLU & - \\
Dropout & 128 & 0 & - & p = 0.5 \\
Linear & 10 & 1{,}290 & Softmax & - \\
\Hline
\end{tabular}
}
\end{table}

\begin{table}[h]
\centering
\caption{Model architectures for CIFAR-10.}
\label{tab:model_cifar}
\scalebox{0.9}{
\begin{tabular}{lcccc}\Hline
Layer & Output Shape & Trainable Parameters & Activation & Hyperparameters \\\hline
Input & (32, 32, 3) & 0 & - & - \\
Conv2d & (32, 32, 64) & 1{,}792 & ReLU & kernel size = 3; strides=(1, 1) \\
Conv2d & (32, 32, 64) & 36{,}928 & ReLU & kernel size = 3; strides=(1, 1) \\
MaxPool2d & (16, 16, 64) & 0 & - & pool size= (2, 2) \\
Conv2d & (16, 16, 128) & 73{,}856 & ReLU & kernel size = 3; strides=(1, 1) \\
Conv2d & (16, 16, 128) & 147{,}584 & ReLU & kernel size = 3; strides=(1, 1) \\
MaxPool2d & (8, 8, 128) & 0 & - & pool size= (2, 2) \\
Conv2d & (8, 8, 256) & 295{,}168 & ReLU & kernel size = 3; strides=(1, 1) \\
Conv2d & (8, 8, 256) & 590{,}080 & ReLU & kernel size = 3; strides=(1, 1) \\
MaxPool2d & (4, 4, 256) & 0 & - & pool size= (2, 2) \\
Conv2d & (4, 4, 512) & 1{,}180{,}160 & ReLU & kernel size = 3; strides=(1, 1) \\
Conv2d & (4, 4, 512) & 2{,}359{,}808 & ReLU & kernel size = 3; strides=(1, 1) \\
MaxPool2d & (2, 2, 512) & 0 & - & pool size= (2, 2) \\
Conv2d & (2, 2, 512) & 2{,}359{,}808 & ReLU & kernel size = 3; strides=(1, 1) \\
Conv2d & (2, 2, 512) & 2{,}359{,}808 & ReLU & kernel size = 3; strides=(1, 1) \\
MaxPool2d & (1, 1, 512) & 0 & - & pool size= (2, 2) \\
Dropout & 512 & 0 & - & p = 0.5 \\
Linear & 512 & 262{,}656 & ReLU & - \\
Dropout & 512 & 0 & - & p = 0.5 \\
Linear & 512 & 262{,}656 & ReLU & - \\
Linear & 10 & 5{,}130 & Softmax & - \\

\Hline
\end{tabular}
}
\end{table}

\section{Hyper-parameters setting}
\label{appendix:hyperparameters}
We show the hyper-parameters that we used in our experimental study.
We generally follow the hyper-parameter settings of original papers that each method has been proposed.

\smallskip\noindent
\textbf{All methods.}
The common hyper-parameters of all methods are listed as follows:
\begin{itemize}
    \item Momentum: $0.9$
    \item Weight decay: $10^{-4}$
    \item Max norm of the gradients: $20$
    \item Batch size
    \begin{itemize}
        \item For FEMNIST: 20
        \item For Shakespeare: 20
        \item For Sent140: 40
        \item For MNIST: 10
        \item For CIFAR-10: 4
    \end{itemize}
    \item Learning rate: See Table~\ref{tab:opt_lr}.
\end{itemize}
We here note that the learning rates were tuned in $\{10^{-3}, 10^{-2.5}, 10^{-2}, 10^{-1.5}, 10^{-1}, 10^{0.5}, 10^{0}, 10^{0.5}\}$ for each dataset and method.

\begin{table}[t]
\centering
\caption{The optimal learning rates.}
\label{tab:opt_lr}
\scalebox{1.0}{
\begin{tabular}{lccccc}\Hline
 & FEMNIST & Shakespeare & Sent140 & MNIST & CIFAR-10 \\\hline
FedAvg & $10^{-2}$ &$10^{-1}$ &$10^{-1}$ & $10^{-2.5}$ & $10^{-2}$ \\
FedProx & $10^{-2}$ &$10^{-0.5}$ &$10^{-1}$ & $10^{-2.5}$ & $10^{-2}$ \\
HypCluster &$10^{-1.5}$ &$10^{-3}$ &$10^{-1.5}$ & $10^{-2.5}$ & $10^{-2}$ \\
FML & $10^{-2}$ &$10^{-0.5}$ &$10^{-0.5}$ & $10^{-2.5}$ & $10^{-2}$ \\
FedMe &$10^{-2}$ &$10^{-1.5}$ & $10^{-2.5}$ &$10^{-3}$ &$10^{-2}$ \\
LG-FedAvg &$10^{-3}$ &$10^{-0.5}$ & $10^{-2.5}$ &$10^{-2}$ &$10^{-2}$ \\
FedPer &$10^{-1.5}$ &$10^{-1.5}$ &$10^{-2}$ & $10^{-2.5}$ &$10^{-2}$ \\
FedRep &$10^{-1.5}$ &$10^{-1}$ &$10^{-1}$ & $10^{-2.5}$ & $10^{-2.5}$ \\
Ditto &$10^{-2}$ &$10^{-0.5}$ & $10^{-2.5}$ &$10^{-3}$ &$10^{-1.5}$ \\
pFedMe &$10^{-2}$ &$10^{-1}$ &$10^{-1.5}$ &$10^{-2}$ & $10^{-2.5}$ \\\hdashline
Local Data Only & $10^{-2.5}$ &$10^{-0.5}$ &$10^{-2}$ & $10^{-2.5}$ &$10^{-2}$ \\
Centralized & $10^{-2.5}$ &$10^{-1}$ &$10^{-2}$ &$10^{-3}$ &$10^{-2}$ \\
\Hline
\end{tabular}
}
\end{table}

The followings are the method-specific hyper-parameters.

\smallskip\noindent
\textbf{FedProx.}
The hyper-parameters of FedProx are listed as follows:
\begin{itemize}
    \item Parameter to control the regularization term $\mu$: $0.001$
\end{itemize}

\smallskip\noindent
\textbf{HypCluster.}
The hyper-parameters of HypCluster are listed as follows:
\begin{itemize}
    \item The number of clusters $k$: $2$
\end{itemize}

\smallskip\noindent
\textbf{FedMe.}
The hyper-parameters of FedMe are listed as follows:
\begin{itemize}
    \item The numbers of global communication rounds to increase the number of clusters
    \begin{itemize}
        \item For FEMNIST: 150, 225, and 275
        \item For Shakespeare: 50, 75, and 90
        \item For Sent140: 25, 50, and 75
        \item For MNIST: 50, 100, and 150
        \item For CIFAR-10: 250, 375, and 450
    \end{itemize}
    \item The number of unlabeled data: 1000
\end{itemize}

\smallskip\noindent
\textbf{LG-FedAvg.}
The hyper-parameters of LG-FedAvg are listed as follows:
\begin{itemize}
    \item Sub-models of the server and the clients
    \begin{itemize}
        \item For FEMNIST, Shakespeare, Sent140, and MNIST
        \begin{description}
            \item[Server: ]The last linear layer
            \item[Clients: ]The all layers except for the last linear layer
        \end{description}
        \item For CIFAR-10
        \begin{description}
            \item[Server: ]The all linear layers
            \item[Clients: ]The all convolutional layers
        \end{description}
    \end{itemize}
\end{itemize}

\smallskip\noindent
\textbf{FedPer.}
The hyper-parameters of FedPer are listed as follows:
\begin{itemize}
    \item Sub-models of the server and the clients
    \begin{itemize}
        \item For FEMNIST, Shakespeare, Sent140, and MNIST
        \begin{description}
            \item[Server: ]The all layers except for the last linear layer
            \item[Clients: ]The last linear layer
        \end{description}
        \item For CIFAR-10
        \begin{description}
            \item[Server: ]The all convolutional layers
            \item[Clients: ]The all linear layers
        \end{description}
    \end{itemize}
\end{itemize}

\smallskip\noindent
\textbf{FedRep.}
The hyper-parameters of FedRep are listed as follows:
\begin{itemize}
    \item The number of epochs to train sub-models: $2$ for each of sub-models
    \item Sub-models of the server and the clients
    \begin{itemize}
        \item For FEMNIST, Shakespeare, Sent140, and MNIST
        \begin{description}
            \item[Server: ]The all layers except for the last linear layer
            \item[Clients: ]The last linear layer
        \end{description}
        \item For CIFAR-10
        \begin{description}
            \item[Server: ]The all convolutional layers
            \item[Clients: ]The all linear layers
        \end{description}
    \end{itemize}
\end{itemize}

\smallskip\noindent
\textbf{Ditto.}
The hyper-parameters of Ditto are listed as follows:
\begin{itemize}
    \item Parameter to control the interpolation between global and personalized models $\lambda$: $0.75$
\end{itemize}

\smallskip\noindent
\textbf{pFedMe.}
The hyper-parameters of pFedMe are listed as follows:
\begin{itemize}
    \item Parameter to control the regularization term $\lambda$: $15$
    \item The number of repetitions of batch trains $K$: $5$
\end{itemize}

\end{document}